\def\year{2021}\relax
\documentclass[letterpaper]{article} 
\usepackage{aaai21}  
\usepackage{times}  
\usepackage{helvet} 
\usepackage{courier}  
\usepackage[hyphens]{url}  
\usepackage{graphicx} 
\usepackage{natbib}  
\usepackage{caption} 
\nocopyright
\usepackage{color} 
\usepackage{amsmath}
\usepackage{amssymb}
\usepackage{amssymb}
\usepackage{pifont}
\newcommand{\cmark}{\ding{51}}%
\newcommand{\xmark}{\ding{55}}%
\newcommand{\fewshot}{\ding{108}}%
\newcommand{\metadata}{\ding{110}}%
\newcommand{\vect}[1]{\boldsymbol{#1}}
\usepackage{array}
\newcolumntype{C}[1]{>{\centering\let\newline\\\arraybackslash\hspace{0pt}}m{#1}}
\usepackage{algorithm}
\usepackage[noend]{algpseudocode}
\usepackage{algorithmicx,algorithm}
\title{Data-Free Knowledge Distillation with Soft Targeted Transfer Set Synthesis}
\author{

    Zi Wang
    \\
}
\affiliations{

    Department of Electrical Engineering and Computer Science, The University of Tennessee\\
    zwang84@vols.utk.edu

}

\begin{document}

\maketitle

\begin{abstract}
Knowledge distillation (KD) has proved to be an effective approach for deep neural network compression, which learns a compact network (student) by transferring the knowledge from a pre-trained, over-parameterized network (teacher). In traditional KD, the transferred knowledge is usually obtained by feeding training samples to the teacher network to obtain the class probabilities. However, the original training dataset is not always available due to storage costs or privacy issues. In this study, we propose a novel data-free KD approach by modeling the intermediate feature space of the teacher with a multivariate normal distribution and leveraging the soft targeted labels generated by the distribution to synthesize pseudo samples as the transfer set. Several student networks trained with these synthesized transfer sets present competitive performance compared to the networks trained with the original training set and other data-free KD approaches.
\end{abstract}

\section{Introduction}
In recent years, deep neural networks (DNNs) have been widely applied to various applications such as object classification and detection \cite{krizhevsky2012imagenet,ren2015faster,huang2017densely}, image synthesis \cite{goodfellow2014generative,gulrajani2017improved,li2018fast}, and robotic control \cite{levine2018learning}. However, as state-of-the-art performance is usually acquired by leveraging deeper and wider architectures \cite{simonyan2014very,szegedy2015going,he2016deep,huang2017densely}, over-parameterization becomes a critical issue that prohibits DNN's usage on resource-efficient platforms such as mobile phones and drones \cite{zhang2018shufflenet}. Experts have been putting great effort into compressing large, cumbersome DNNs with the approaches such as quantization \cite{gong2014compressing,han2015deep}, channel pruning \cite{li2016pruning,wang2019pruning,wang2019towards,you2020greedynas}, and knowledge distillation (KD) \cite{hinton2015distilling,liu2019exploiting,heo2019knowledge,jin2020uncertainty}. 

Among all these approaches, KD is a popular scheme that trains a smaller model (student) to mimic the softmax outputs of a pre-trained over-parameterized model (teacher) \cite{hinton2015distilling}. With this approach, the performance of the student model can be improved compared to training the model solely with the cross-entropy loss. However, classic KD methods usually rely on the dataset that contains labeled samples to transfer the knowledge from the teacher to the student, which is a strong constraint because the training set is not always available due to the following reasons. (1) Most importantly, the dataset used for training the teacher model is not often publicly shared because of the concerns of conflict of interest, privacy issues, or business competition \cite{taigman2014deepface, wu2016google}. (2) SOTA models are usually trained with extremely large datasets. For example, the popular object classification dataset ImageNet \cite{deng2009imagenet} contains more than one million training samples that needs more than 100 GB of storage space. Spreading such a large dataset frequently across devices or on the Internet is a heavy burden and a waste of resources.

Data-free KD, or zero-shot KD, was first introduced in \cite{nayak2019zero} to deal with the problem that labeled training samples are missing. Starting from some targets (output probabilities) obtained with certain prior knowledge, noise inputs are optimized to minimize the distances between the softmax outputs obtained by feeding these noise inputs to the teacher network and the targets. Finally, the optimized samples are used for training the student network via a standard KD procedure. The key idea for the implementation of data-free KD is to generate informative pseudo samples that can capture the distribution of the original training samples. For example, \cite{nayak2019zero} models the softmax space of the teacher network as a Dirichlet distribution and generates images by optimizing the noise input to mimic the softmax outputs sampled from the distribution. Generative adversarial networks \cite{goodfellow2014generative} are used in \cite{chen2019data} for derivating training samples. By considering the pre-trained teacher network as a fixed discriminator, the generator aims to generate samples that cause the maximum response on the discriminator. Although a few studies have been proposed \cite{nayak2019zero,chen2019data,micaelli2019zero}, KD in the absence of prior training data is still not well studied and there are clear opportunities to improve on the performance of existing approaches.

In this paper, we present a novel data-free KD approach as an enrichment of this new line of research. Specifically, we propose to model the output of an intermediate layer of the teacher model with a multivariate normal distribution and obtain the soft targets with the outputs sampling from the distribution. Then the soft targets are used for training sample synthesis by optimizing noise inputs via backpropagation. Finally, the optimized samples are used to train the student network via a standard KD procedure. Different from existing works that directly model the softmax space to obtain the targets, we argue that modeling shallower feature spaces, and feeding the generated intermediate feature representations to the following layers to obtain soft targets helps improve the performance. As features transition from general to specific when the data flow to deeper layers \cite{yosinski2014transferable}, modeling the output distribution of shallower layers can obtain more generalized soft targets compared to modeling the softmax space directly.

We summarize the main contributions of our study as follows.
\begin{itemize}
    \item We model the feature space of the teacher's intermediate layer with a multivariate normal distribution and optimize pseudo samples towards the targets sampled from the distribution. By doing so, the quality of the synthesized samples is improved, which helps train the student better.
    \item We model the output distribution of the shallower layer, rather than directly modeling the softmax space for targets sampling, so that more generalized soft targets can be obtained, which helps improve the performance.
    \item The proposed approach is evaluated with various benchmark network architectures and datasets and exhibits clear improvement over existing works. Specifically, our student networks trained with the proposed approach achieve 99.08\% and 93.31\% accuracies without using any original training samples by transferring the knowledge from the teacher networks pre-trained on the MNIST and CIFAR-10 datasets.
\end{itemize}

\section{Related Work}
\label{sec:work}
\subsection{Traditional Knowledge Distillation}
The idea of KD was initially proposed by \cite{bucilua2006model} and was substantially developed by \cite{ba2014deep} in the era of deep learning. It trains a smaller student network by matching the logits (also called log probability values) before the softmax activations obtained from a cumbersome network. \cite{hinton2015distilling} extended this idea by softening the softmax output with a scaling factor called temperature, which produces knowledge with higher entropy and improves the performance of the student network. By doing so, \cite{hinton2015distilling} becomes a generalized case of \cite{ba2014deep}. Recently, a number of variants have been proposed by adding extra regulations/alignments to the vanilla KD approach. For example, FitNets \cite{romero2014fitnets} uses $\ell_2$-norm to map the intermediate feature representations of the student with the pre-trained teacher so that a deeper and thinner student than the teacher can be well trained. Attention maps are calculated from the teacher's intermediate feature representations in \cite{zagoruyko2016paying} as an extra alignment for knowledge transfer. MEAL \cite{shen2019meal} proposes to distill the knowledge from multiple teachers via adversarial learning.  

\subsection{Few-Shot and Meta-Data Knowledge Distillation}
Due to the storage costs of large scale datasets, efficient KD approaches using limited training samples are investigated by several studies. In \cite{kimura2018few}, the authors first trained a reference model via KD with only a limited amount of training samples. Then a data augmentation approach is proposed to help increase the performance of the student model by generating pseudo samples via the inducing point method \cite{snelson2006sparse}. \cite{ahn2019variational} proposes variational information distillation (VID), which aims to maximize the mutual information between the teacher and the student models for few-shot KD. $1 \times 1$ convolution layers are added at the end of all the layer blocks of the student model in \cite{li2020few}. By matching the block-level outputs of the teacher and the student models, distilling the knowledge to the student model can be achieved with only a few label-free samples.

Instead of transferring knowledge with labeled samples, alternatives such as meta-data are also used for generating pseudo samples to train the student model. \cite{lopes2017data} stores the activation records of certain layers when training the teacher model as the meta-data and uses them to reconstruct the original training samples by optimizing noise inputs, which are then used as the transfer set for training the student. However, releasing such kinds of meta-data along with the pre-trained network is an unusual scenario for most of the applications.

As mentioned before, all the above KD approaches rely on either the labeled data or their surrogates to produce the class probabilities as the matching targets, which are not always available in practice.

\subsection{Data-Free Knowledge Distillation}
Data-free KD, or zero-shot KD (ZSKD), was first introduced in \cite{nayak2019zero}, which transfers the knowledge from the teacher to the student without any type of prior information of the training set. ZSKD \cite{nayak2019zero} models the class probabilities with a Dirichlet distribution and generate labels from the distribution to obtain pseudo training samples. Data-Free Learning (DAFL) \cite{chen2019data} considers the teacher model as a fixed discriminator and trains a generator to generate images that can produce similar softmax outputs from the teacher and the student model. The student model is trained simultaneously with the generator via KD. Adversarial Belief Matching (ABM) was proposed in \cite{micaelli2019zero}, which trains a generative adversarial network \cite{goodfellow2014generative} to search for samples on which the student model poorly matches the teacher, and then train the student with the generated samples. DeepInversion \cite{yin2020dreaming} takes the information stored in the batch normalization layers of the teacher to synthesize images that are used as the transfer set.

Our proposed approach is related to ZSKD, but differs from it in the following ways. (1) ZSKD generates soft targeted labels for each specific class by modeling the softmax space with a Dirichlet distribution. Due to the inherent property of Dirichlet distribution, labels that are mismatched with their real categories are produced (see the ablation study and analysis section for details). We consider all the classes as a whole and use a multivariate normal distribution to model the feature space to resolve this problem. (2) ZSKD models the distribution of the softmax space, however, we argue that modeling the data in the shallower feature space generalizes the feature representation better and can therefore improve the performance. (3) We use an extra activation loss term to encourage higher activation values during the image synthesis procedure, which is not used in ZSKD.

\section{The Proposed Approach}
\label{sec:approach}
In this section, we first briefly present the procedure of standard KD. Then we introduce our proposed data-free KD approach from the following aspects. (1) Generating soft targeted labels by modeling the intermediate feature space with a multivariate normal distribution. (2) Generating pseudo training samples by optimizing the noise inputs towards the generated soft targeted labels with the pre-trained, fixed teacher model. (3) Using the generated samples as the transfer set to train the student model via standard KD.

\subsection{Knowledge Distillation}
Knowledge distillation \cite{hinton2015distilling} is a popular model compression approach by training a compact student model ($S$) to mimic the softmax outputs of a pre-trained cumbersome teacher model ($T$). Let $W_T$ and $W_S$ be the parameters of the teacher and the student model, respectively. The softmax outputs (class probabilities) of the teacher and the student model are represented as $P_T=T(x,W_T)=\text{softmax}(a_T)$ and $P_S=T(x,W_S)=\text{softmax}(a_S)$, respectively, where $x$ is the training sample and $a$ is the pre-softmax activation of a model. During the KD process, a temperature $\tau$ is usually used for softening the class probability (Eq. \eqref{eq:temperature}). A larger $\tau$ can produce softer class probabilities so that information with higher entropy is enclosed in the targets, and the student model can be trained with a larger learning rate and converge faster.
\begin{equation}
\begin{aligned}
&P^\tau_T = T(x,W_T,\tau)=\text{softmax}(\frac{a_T}{\tau}),\\
&P^\tau_S = T(x,W_S,\tau)=\text{softmax}(\frac{a_S}{\tau}).
\end{aligned}
\label{eq:temperature}
\end{equation}

$W_S$ can be learned by minimizing the loss function in Eq. \eqref{eq:vanillakd}.
\begin{equation}
\mathcal{L}_{KD}=\mathcal{L}_{CE}(P^\tau_T,P^\tau_S)+\lambda_c\mathcal{L}_{CE}(P_S,y),
\label{eq:vanillakd}
\end{equation}
where $y$ is the one-hot ground truth vector, $\mathcal{L}_{CE}(\cdot)$ is the cross-entropy loss, and $\lambda_c$ is a scaling factor that balances the importance of the two losses.

\subsection{Data-Free Knowledge Distillation with Soft Targeted Transfer Set Synthesis}
In the absence of the original training dataset, pseudo samples have to be generated as the carrier for knowledge transfer. In this study, we propose to model the intermediate feature representation of the teacher model with a multivariate normal distribution. We then use this distribution to generate samples as the feature representations, which are used as either the soft targeted labels, or for producing the soft targeted labels by feeding these samples to the rest layers of the teacher model, depending on whether the softmax space or the intermediate feature space is modeled. We then generate pseudo samples by optimizing the noise inputs through backpropagation with the fixed teacher model. This is achieved by minimizing the distances between the softmax outputs corresponding to the inputs to be optimized and the generated soft targets. Finally, the student model is trained with the pseudo samples through a standard KD process.

\subsubsection{Feature space modeling with multivariate normal distribution}
Suppose the pre-trained teacher model has $L$ layers parameterized with $W_T=\{W_T^1, W_T^2,\cdots, W_T^L\}$. We use a $k$-dimensional multivariate random variable $\vect{s^l}=\{s_1^l,s_2^l,\cdots,s_k^l\}\sim p(\vect{s^l})$ to represent the output space of the $l$-th layer of the teacher model ($l=1,2,\cdots,L$), where $s_i^l$ is the $i$-th element, and $k$ is the number of elements in the feature space. We model $\vect{s^l}$ as a multivariate normal distribution, i.e., $\vect{s^l}\sim \mathcal{N}_k(\vect{\mu},\vect{\Sigma})$, where $\vect{\mu}\in \mathbb{R}^{k}$ is the mean vector and $\vect{\Sigma} \in \mathbb{R}^{k \times k}$ is the covariance matrix, respectively. 

Statistically, $\vect{\Sigma}=(\sigma_{ij})_{k\times k}$ can be obtained from its corresponding correlation matrix $\vect{R}=(\rho_{ij})_{k \times k}$. Let $\vect{D}=\text{diag}[\sqrt{\sigma_{11}}, \sqrt{\sigma_{22}}, \cdots, \sqrt{\sigma_{kk}}]$, then 
\begin{equation}
\begin{aligned}
\Sigma = \vect{D} \times \vect{R} \times \vect{D},
\end{aligned}
\label{eq:covariance}
\end{equation}
or,
\begin{equation}
\begin{aligned}
\sigma_{ij} = \rho_{ij} \sqrt{ \sigma_{ii} \sigma_{jj}}, \ ~~~i,j = 1,2,\cdots, k.
\end{aligned}
\end{equation}

Compared to the covariance matrix $\vect{\Sigma}$, the coefficient matrix $\vect{R}$ is a more intuitive statistic that represents the correlation among different components in the random vector, which can be intuitively obtained from the weights of the teacher model. For simplicity, here we consider modeling the feature spaces of the fully connected layers in the teacher model. Suppose we model the feature space of the $l$-th layer in the teacher model, inspired by \cite{nayak2019zero}, we claim that the weights that are used for calculating the $l$-th layer's feature maps can be considered as a template learned by the teacher model. This template aligns the neurons of the $(l-1)$-th layer to the neurons of the $l$-th layer and maps the relationship between the feature maps of these two layers. If the value of an element in the feature space peaks, it means that the corresponding weights activate it. On the other hand, if the relationship is misaligned, the value of the certain element decreases. Therefore, we claim that the correlation between the elements of a layer's feature space is implicitly hidden in the weights of this layer, which can be calculated with Eq. \eqref{eq:similarity}.
\begin{equation}
R(i,j)=\rho_{ij}=\frac{\vect{w_i}^T\vect{w_j}}{||\vect{w_i}||\cdot||\vect{w_j}||},
\label{eq:similarity}
\end{equation}
where $\vect{w_i}$ denotes the weights connecting the $(l-1)$-th layer to the $i$-th element in the $l$-th layer.

Determining the variance of each variable $\sigma_{ii}~(i=1,2,\cdots,k)$ in $\vect{D}$ is not straightforward in the data-free KD scenario. Therefore, we consider $\vect{D}$ as a hyperparameter.
Actually, $\sigma$ can be considered as a concentration factor that controls the density of the feature space. A larger $\sigma$ leads to a distribution that concentrated on one or a few components in the feature representations. On the other hand, when $\sigma$ gets smaller, the samples are closer to a uniform distribution (see ablation study for details). Moreover, we empirically find that changing the value of $\vect{\mu}$ only leads to trivial difference of performance. A reasonable explanation on this phenomenon is that $\vect{\mu}$ only shifts the center of the feature space, which can largely be canceled out by the softmax operation in the last layer. Therefore, in our empirical studies, we just use $\vect{\mu}=\vect{0}$ for simplicity.

\subsubsection{Soft targeted label generation} 
Once the feature space of the $l$-th layer is modeled with a multivariate distribution, and the values of $\vect{R}$, $\vect{D}$, and $\vect{\mu}$ are obtained, we can calculate $\vect{\Sigma}$ with Eq. \eqref{eq:covariance} and generate samples $\vect{s^l}$ from the distribution as the $l$-th layer's outputs of the teacher model. If the last layer (softmax space) is modeled, the soft targeted label $\vect{y^{\text{soft}}}$ can be obtained with Eq. \eqref{eq:label_last}.
\begin{equation}
\vect{y^\text{soft}}=\text{softmax}(\frac{\vect{s^l}}{\tau}).
\label{eq:label_last}
\end{equation}

Otherwise, denote $W_T^{l+}$ the weights after the $l$-th layer of the teacher model. The soft targeted label can be obtained by feeding $\vect{s^l}$ to the rest layers of the teacher model (Eq. \eqref{eq:label_notlast}).
\begin{equation}
\vect{y^\text{soft}}=T(\vect{s^l},W_T^{l+},\tau).
\label{eq:label_notlast}
\end{equation}

We argue that modeling the feature space of a shallower layer can improve the performance, compared to modeling the softmax space \cite{nayak2019zero}. Since features gradually transition from general to specific as the tensors feed forward to deeper layers \cite{yosinski2014transferable}, modeling the feature space of a shallower layer will boost generalization (see ablation study for details).

\subsubsection{Sample generation}
We then use the generated soft labels as the targets to produce pseudo training samples. We first randomly sample $n$ soft targets $\vect{y_i^\text{soft}}~ (i=1,2,\cdots,n)$ and generate a batch of $n$ noise inputs $\hat{x}_i~(i=1,2,\cdots,n)$. By feeding $\hat{x}_i$ into the fixed teacher model, we obtain the corresponding labels $\vect{\hat{y}_i}$. The noise inputs are optimized with an iterative backpropagation process to minimize the distance between $\vect{y_i^\text{soft}}$ and $\vect{\hat{y}_i}$. There are various criteria that can be utilized to minimize this distance and we choose to use the Kullback–Leibler (KL) divergence (Eq. \eqref{eq:distmin}).
\begin{equation}
\mathcal{L}_{d} = \mathcal{L}_{KL}(\vect{y_i^\text{soft}}, \vect{\hat{y}_i}),
\label{eq:distmin}
\end{equation}
where $\mathcal{L}_{KL}(\cdot)$ is the KL divergence, and $\vect{\hat{y}_i}=T(\hat{x}_i,W_T,\tau)$.

It has been proved that a well trained deep neural network usually receives higher activation values when the training samples are fed. Therefore, we define an extra activation loss $\mathcal{L}_{a}$ to encourage higher activation values of the last convolutional layer $x_{\text{conv-1}}$ during the image synthesis procedure (Eq. \eqref{eq:activation}).
\begin{equation}
\mathcal{L}_{a} = -\frac{1}{n}\Sigma_{i=1}^n||x_{\text{conv-1}}^i||_1.
\label{eq:activation}
\end{equation}

Therefore, the total loss of the sample generation process is defined as Eq. \eqref{eq:total_gen}.
\begin{equation}
\mathcal{L}_{sg} = \mathcal{L}_{d} + \lambda_a\mathcal{L}_{a},
\label{eq:total_gen}
\end{equation}
where $\lambda_a$ is a scaling factor that balances the importance of the two terms.

\subsubsection{Knowledge distillation with pseudo samples}
Finally, we use the generated pseudo samples as the transfer set to train the student model with Eq. \eqref{eq:zskd}.
\begin{equation}
\mathcal{L}_{DFKD}=\mathcal{L}_{CE}(T(\hat{x},W_T,\tau),S(\hat{x},W_S,\tau)).
\label{eq:zskd}
\end{equation}
It is worth mentioning that here we omit the second term compared to Eq. \eqref{eq:vanillakd} when using the transfer set to train the student model. This is because pseudo samples, rather than real training samples, are used for the KD process. Adding a cross-entropy loss with one-hot labels produces little extra meaningful information in this scenario.

We summarize our proposed approach in Algorithm 1.
\begin{algorithm}[t]
\caption{Data-free knowledge distillation for compact student model training}
\label{alg}
{\bf Input:} 
The teacher $T$ with $L$ layers parameterized with $W_T$, the index of the layer $l$ whose output space is modeled, the temperature $\tau$, the activation loss scaling factor $\lambda_a$, the mean $\vect{\mu}$ and variance $\vect{D}$ of the elements in the modeled feature space, the number of training iteration $N$, the batch size $n$.

{\bf Output:} The learned student model $S$.
\begin{algorithmic}[1]
\State Initialize the transfer set $\hat{X}=\emptyset$
\State Compute the coefficient matrix $\vect{R}$ with Eq. \eqref{eq:similarity}
\State Compute the covariance matrix $\vect{\Sigma}$ with Eq. \eqref{eq:covariance}
\For {$i$ in range(N)}
\State Generate $n$ samples: $\vect{s^l_n} \sim \mathcal{N}(\vect{\mu},\vect{\Sigma})$
\If {$l==L$}
\State $\vect{y^{\text{soft}}_n}=\text{softmax}(\frac{\vect{s^l_n}}{\tau})$
\Else
\State $\vect{y^\text{soft}_n}=T(\vect{s^l_n},W_T^{l+},\tau)$
\EndIf
\State Randomly initialize $\hat{x}_n$ and optimize it with Eq. \eqref{eq:total_gen}
\State $\hat{X} \leftarrow \hat{X} \cup \hat{x}_n$
\EndFor
\State Using $\hat{X}$ to train the student model via KD with Eq. \eqref{eq:zskd}\\
\Return $S$
\end{algorithmic}
\end{algorithm}

\section{Experiments}
\label{sec:exp}

\subsection{Setup}
We evaluate our proposed data-free KD approach on object classification tasks with the following configurations. (1) A LeNet-5 \cite{lecun1998gradient} pre-trained with the MNIST dataset \cite{lecun1998gradient} following the settings in \cite{lopes2017data,chen2019data} is used as the teacher network. The student model is a LeNet-5-HALF model which contains half the number of filters in each convolutional layer. (2) An AlexNet \cite{krizhevsky2012imagenet} pre-trained with CIFAR-10 \cite{krizhevsky2009learning} as the teacher model and an AlexNet-HALF taking half convolutional filters per layer as the student model. (3) A ResNet-34 \cite{he2016deep} pre-trained with CIFAR-10 as the teacher model and A ResNet-18 as the student model. Architecture details are presented in the Appendix. For each configuration, we first train the teacher model with the cross-entropy loss using a stochastic gradient descent (SGD) optimizer with a batch size of 512 for 200 epochs. The initial learning rate is 0.1, which is divided by 10 at epoch 50, 100, and 150, respectively.

We choose to model the feature space of the second last fully connected layer (represented as $\text{FC}_{-2}$ in the following text) of the teacher model and feed the vectors sampled from the distribution to the last layer to get the soft targeted labels. We also implement experiments that model the softmax space for the purpose of performance comparison. We assume the variances for all the elements in the feature space are the same, i.e., $\sigma=\sigma_{11}=\sigma_{11}=\cdots=\sigma_{kk}$ and implement a hyperparameter search to find the optimal value of $\sigma$. For transfer set synthesis, we generate a batch of 100 soft labels and noise inputs each time. We optimize the noise inputs by minimizing the KL-divergence between their corresponding softmax outputs with the generated labels with an Adam optimizer \cite{kingma2014adam} with a learning rate of 0.001 for 1500 iterations. For the activation loss scaling factor $\lambda_a$, we implement a hyperparameter search and report the best performance with $\lambda_a=0.05, 0.05, 0.1$ for the LeNet-5, AlexNet, and ResNet experiments, respectively. We implement data augmentation on the generated samples, and the transformation includes random rotation, padding and random cropping, scaling, translation, and noise adding. All the student models are trained for 2000 epochs with an Adam optimizer (batch size 512, learning rate 0.001) through a standard KD approach. A temperature ($\tau$) of 20 is used for both of the sample synthesis and student model training across all architectures.

All the experiments are implemented with Tensorflow \cite{abadi2016tensorflow} on an NVIDIA GeForce RTX 2080 Ti GPU and an Intel(R) Core(TM) i7-9700K CPU @ 3.60GHz.

\subsection{Experiments on LeNet-5 with MNIST}
MNIST is a handwritten digits dataset, which contains 60,000 training samples and 10,000 test samples with a resolution of 28x28. In our experiments, the samples are resized to 32x32. The results of LeNet-5 with MNIST is reported in Table \ref{tab:lenet}. Note that training the teacher and the student models with standard cross-entropy loss gives us accuracies of 99.32\% and 98.99\%, respectively. The accuracy of training the student model with a standard KD approach \cite{hinton2015distilling} is 99.18\%, which can be considered as a performance upper bound for data-free KD approaches. Our proposed approach achieves an accuracy of 99.08\%, which is very close to the upper bound and outperforms recent data-dependent (few-shot and meta-data-based) approaches \cite{kimura2018few,lopes2017data} with a clear margin. Compared with other data-free KD approaches, our approach also shows competitive performance, which outperforms the previous state-of-the-art ZSKD by 0.31\%. It's worth noting that using noise images sampling from a standard normal distribution without any optimization, the accuracy is only $87.58\%$, which demonstrates that the transfer set generated with the proposed approach precisely captures the distribution of the original training set.
\begin{table}[t]
\centering
\begin{tabular}{|c|c|c|}
\hline
Model & Data & Accuracy \\
\hline
Teacher (standard training) & \cmark & 99.32\% \\
\hline
Student (standard training) & \cmark & 98.99\% \\
\hline
Standard KD$^*$ & \cmark & 99.18\% \\
\hline
\cite{kimura2018few} & \fewshot & 86.70\% \\
\hline
\cite{lopes2017data} & \metadata & 92.47\% \\
\hline
Noise input & \xmark & 87.58\% \\
\hline
DAFL & \xmark & 98.20\% \\
\hline
ZSKD & \xmark & 98.77\% \\
\hline
Ours & \xmark & 99.08\% \\
\hline
\end{tabular}
\caption{Result on LeNet-5 with the MNIST dataset. Symbols in the ``Data'' column: \cmark: original training data required, \xmark: no data required, \fewshot: limited amount of original training data required (few-shot), \metadata: meta-data required. $^*$ indicates the reported results are based on our own implementation.}
\label{tab:lenet}
\end{table}

\begin{table}[t]
\centering
\begin{tabular}{|c|c|c|}
\hline
Model & Data & Accuracy \\
\hline
Teacher (standard training) & \cmark & 78.56\% \\
\hline
Student (standard training) & \cmark & 75.29\% \\
\hline
Standard KD$^*$ & \cmark & 76.88\% \\
\hline
Noise input & \xmark & 36.49\% \\
\hline
DAFL$^*$ & \xmark & 70.23\% \\
\hline
ZSKD & \xmark & 69.56\% \\
\hline
Ours & \xmark & 73.91\% \\
\hline
\end{tabular}
\caption{Result on AlexNet with the CIFAR-10 dataset. Symbols in the ``Data'' column and $^*$ have the same meanings as in Table \ref{tab:lenet}.}
\vskip -0.2in
\label{tab:alexnet}
\end{table}

\subsection{Experiments on AlexNet with CIFAR-10}
We then evaluate our approach with a more challenging dataset, CIFAR-10 on the AlexNet architecture. The CIFAR-10 dataset consists of 50,000 training samples and 10,000 test samples, which are 32x32 color images of common daily-life objects in 10 classes. Because the vanilla AlexNet \cite{krizhevsky2012imagenet} was designed for the large scale dataset ImageNet \cite{deng2009imagenet}, which takes 227x227 images. We follow \cite{nayak2019zero} to modify the architecture to fit 32x32 inputs. For the first convolutional layer, a 5x5 kernel with a stride of 1 is used. A batch normalization layer is added after each convolutional layer. The modified network contains three fully connected layers, with 512, 256, and 10 neurons, respectively.

The results of AlexNet with CIFAR-10 are presented in Table \ref{tab:alexnet}. It can be observed that our proposed approach that generates soft targeted labels with a multivariate normal distribution achieves an accuracy of 73.91\%, which is the best among all data-free KD approaches. Specifically, our approach outperforms DAFL \cite{chen2019data} and ZSKD \cite{nayak2019zero} by 3.68\% and 4.35\% on the test accuracy, respectively. Since the underlying distribution of the CIFAR-10 training samples is much more complex than the distribution of MNIST, using noise inputs as the transfer set presents a much worse performance on training the student model (36.49\%). These results illustrate the effectiveness of our approach on AlexNet with the CIFAR-10 dataset.

\subsection{Experiments on ResNet with CIFAR-10}
We use ResNets to further evaluate our proposed approach in this sub-section. In this experiment, we use a ResNet-34 pre-trained with CIFAR-10 as the teacher model and a ResNet-18 as the student. Similar to the vanilla AlexNet architecture, ResNet was originally designed for the ImageNet dataset, whose inputs are with the resolution of 224x224. We make the following modifications for the CIFAR-10 dataset: (1) we reduce one convolutional layer (and its following batch normalization layer) before the first residual block, (2) the kernel size and stride of the first convolutional layer are changed to 3 and 1, respectively, (3) the kernel size of the average pooling layer after the last residual block is changed to 4 in order to produce 1x1 feature maps.

The performance comparison of the ResNet experiment is reported in Table \ref{tab:resnet}. Our proposed approach brings the test accuracy quite close to the performance obtained by a standard KD procedure with the original training set (93.31\% vs. 94.40\%). Again, the proposed approach achieves the best performance compared to other data-free KD approaches. It is worth mentioning that our approach achieves similar performance compared to DeepInversion. DeepInversion leverages the statistics in the batch normalization (BN) layer to generate pseudo samples, which can only deal with the scenario in which the teacher has BN layers. Our approach models the intermediate feature space, which is a more generalized case that works for all kinds of networks. Similar to the previous experiments, it can be observed that all the data-free approaches significantly outperform the performance using noise inputs without any optimization to train the student model, which indicates that recovering the prior distribution of the original training samples plays an essential role to achieve good performance.
\begin{table}[t]
\centering
\begin{tabular}{|c|c|c|}
\hline
Model & Data & Accuracy \\
\hline
Teacher (standard training) & \cmark & 95.48\% \\
\hline
Student (standard training) & \cmark & 93.76\% \\
\hline
Standard KD$^*$ & \cmark & 94.40\% \\
\hline
Noise input & \xmark & 19.49\% \\
\hline
DAFL & \xmark & 92.22\% \\
\hline
ZSKD$^*$ & \xmark & 91.99\% \\
\hline
DeepInversion & \xmark & 93.26\% \\
\hline
Ours & \xmark & 93.31\% \\
\hline
\end{tabular}
\caption{Result on ResNet with the CIFAR-10 dataset. Symbols in the ``Data'' column and $^*$ have the same meanings as in Table \ref{tab:lenet}.}
\label{tab:resnet}
\vskip -0.2in
\end{table}

\section{Ablation Study and Analysis}
\label{sec:ana}
\subsection{Performance with Different $\sigma$s}
\begin{figure}[t]
\centering
\includegraphics[width=0.48\columnwidth]{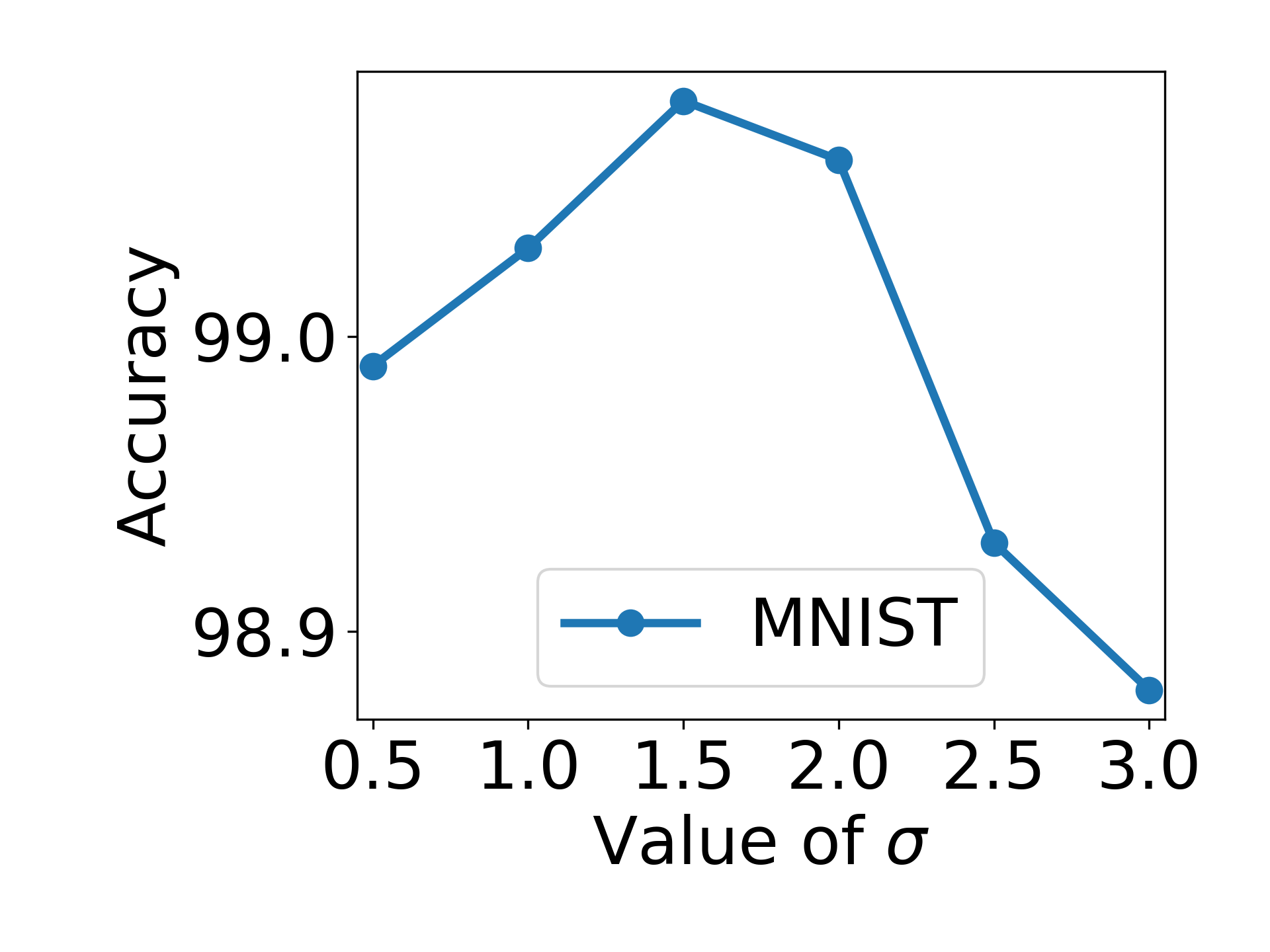}
\includegraphics[width=0.48\columnwidth]{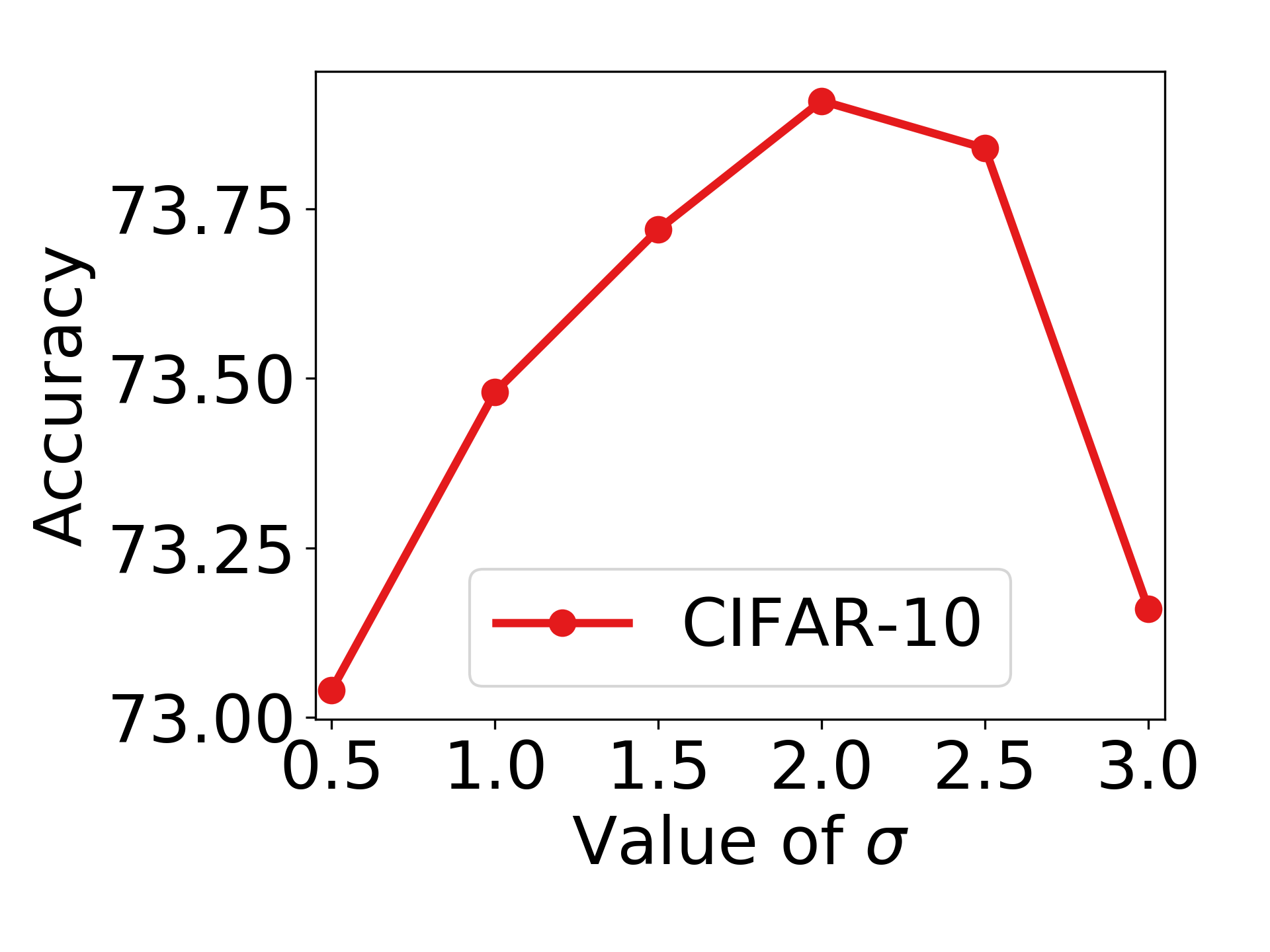}
\vskip -0.1in
\caption{Performance comparison of different choices of $\sigma$. Left: LeNet-5-HALF on MNIST. Right: AlexNet-HALF on CIFAR-10.}
\vskip -0.2in
\label{fig:sigma}
\end{figure}

We first investigate whether different choices of $\sigma$ have an impact on the performance, with LeNet-5-HALF on MNIST and AlexNet-HALF on CIFAR-10. To test this, we fix $\lambda_a=0.05$ and generate the transfer set used for training the student model with $\sigma=0.5, 1.0, 1.5, 2.0, 2.5, 3.0$, respectively. The results are reported in Fig. \ref{fig:sigma}. It appears that the best test accuracy can be obtained when $\sigma=1.5$ and $2.0$ for MNIST and CIFAR-10, respectively. As can be seen from the results, experiments on both configurations exhibit a wide tolerance range of $\sigma$ choices to achieve good performance. With the $\sigma$ values from 0.5 to 3.0, all the trained student models present competitive test accuracy compared to previous state-of-the-art \cite{nayak2019zero}.

We further investigate the influence of $\sigma$ selection by plotting the mean maximal class probabilities (i.e, the maximal value in the soft targets) from the soft targets generated with different $\sigma$s. It is observed that for both configurations, the maximal class probability in the target and its corresponding standard deviation increases as the value of $\sigma$ grows, which indicates that the targets are more concentrated in one or a few components. A too small $\sigma$ generates targets in which all the components look similar, which makes them difficult to be distinguished. On the other hand, a too large $\sigma$ generates targets that are highly concentrated, which leads to less diversity of the soft targets and lower entropy. Both of these two kinds of targets hurt the quality of the generated images. These empirical studies are consistent with the theoretical analysis in the previous section.
\begin{figure}[t]
\centering
\includegraphics[width=0.49\columnwidth]{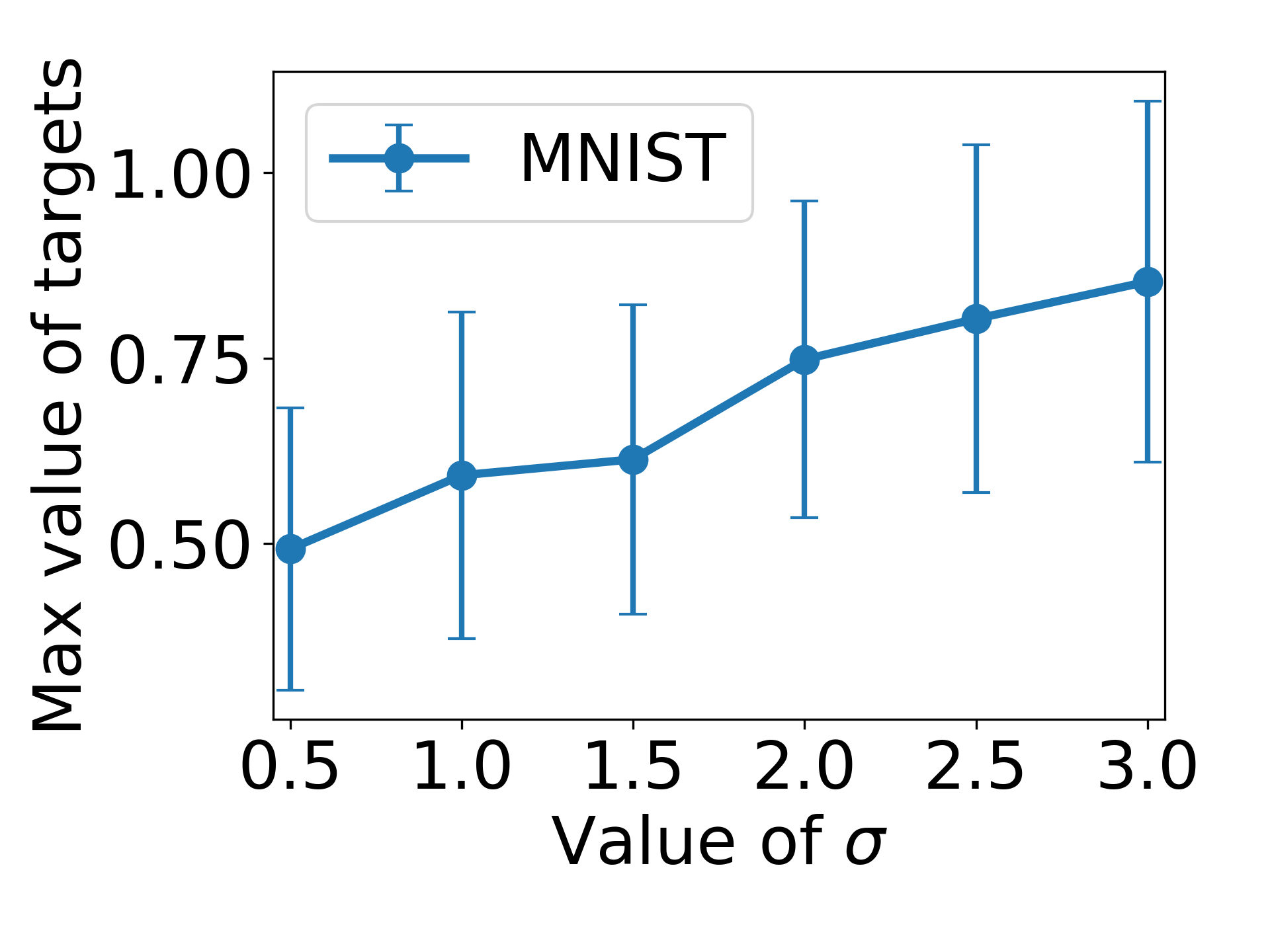}
\includegraphics[width=0.49\columnwidth]{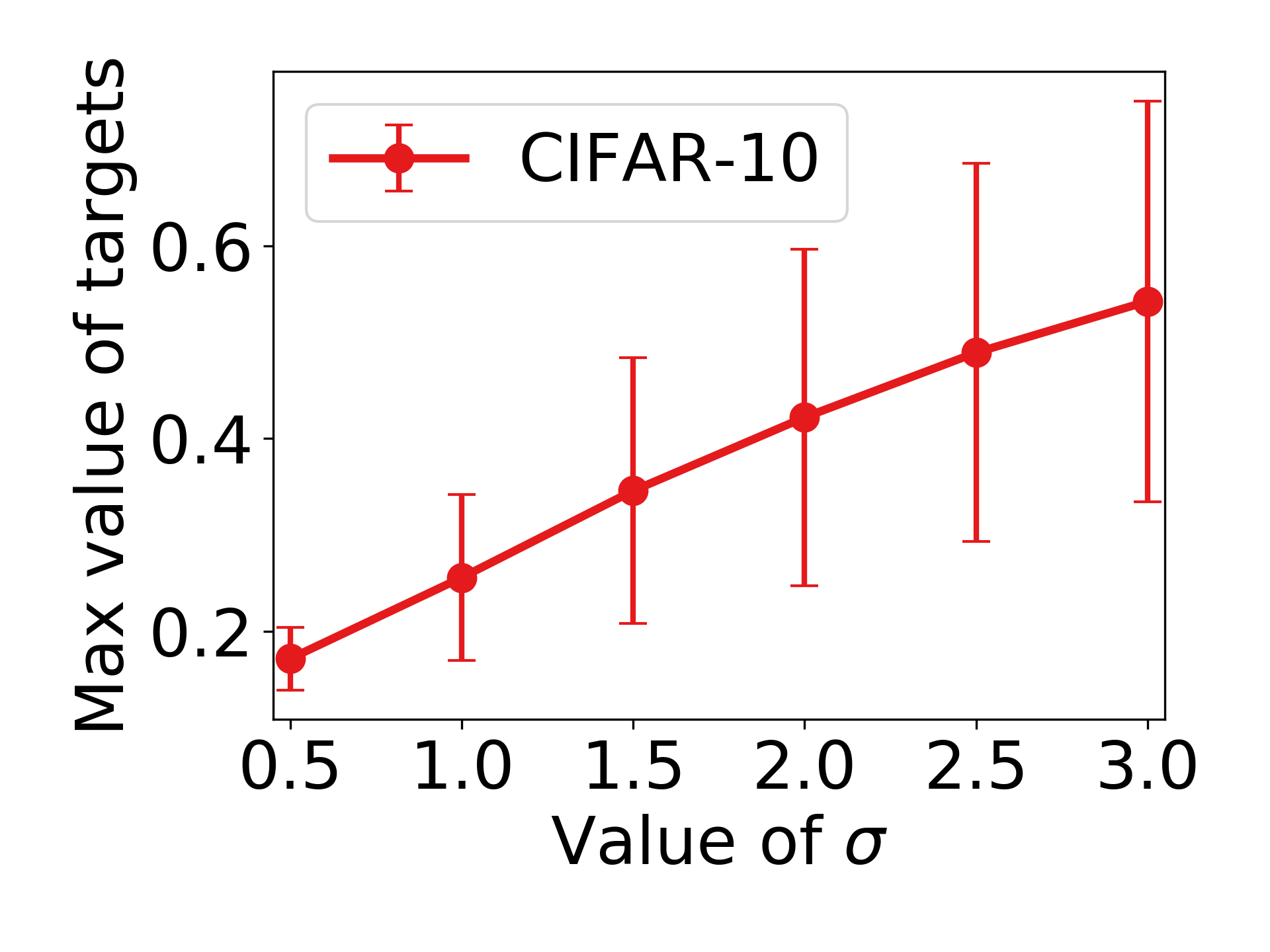}
\vskip -0.1in
\caption{Maximal class probabilities in the soft targets over different $\sigma$s. Left: LeNet-5-HALF on MNIST. Right: AlexNet-HALF on CIFAR-10. Error bar represents the standard deviation.}
\label{fig:maxprob}
\vskip -0.2in
\end{figure}

\subsection{Performance with Different $\lambda_a$}
\begin{figure}[t]
\centering
\includegraphics[width=0.49\columnwidth]{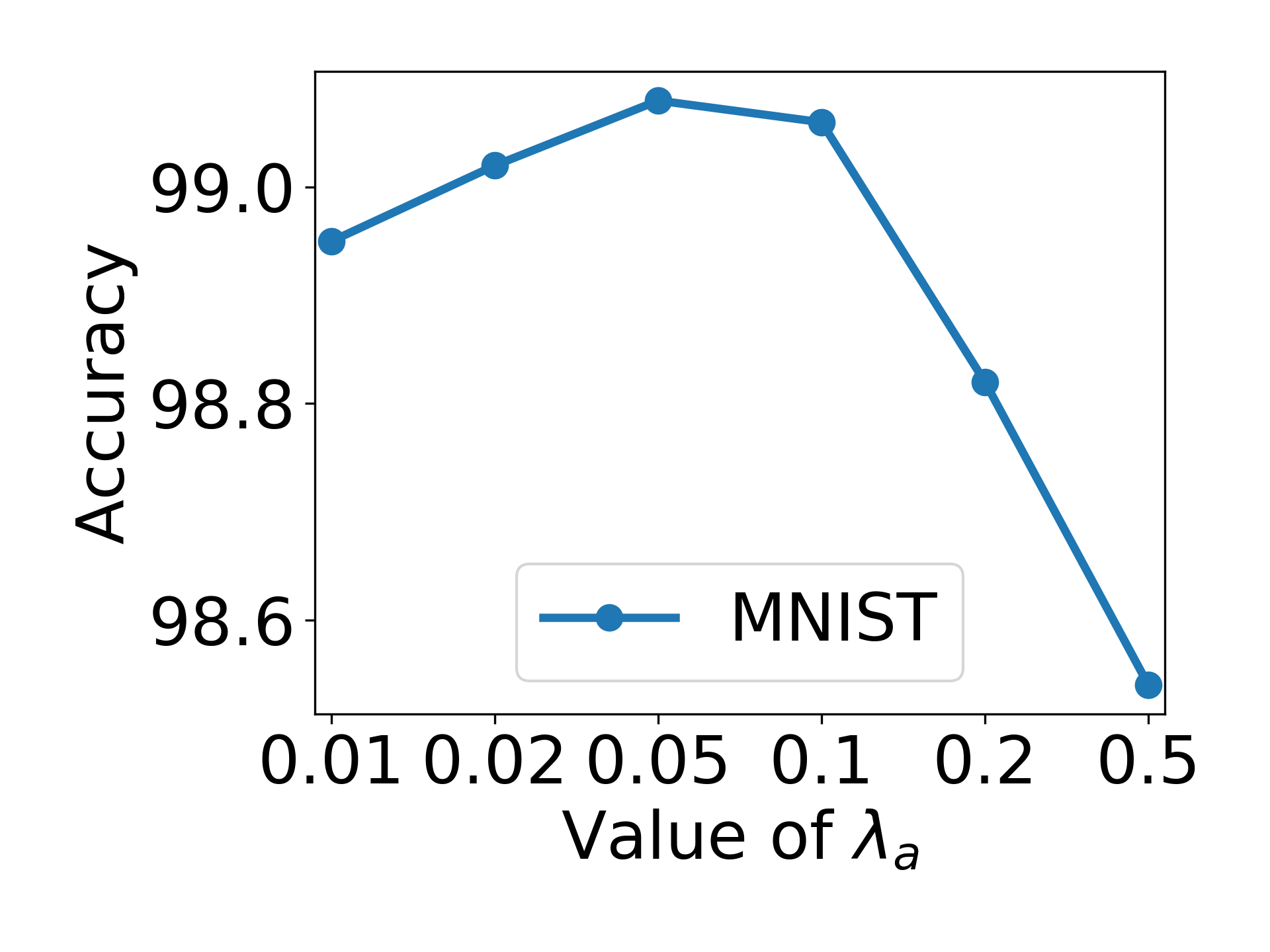}
\includegraphics[width=0.49\columnwidth]{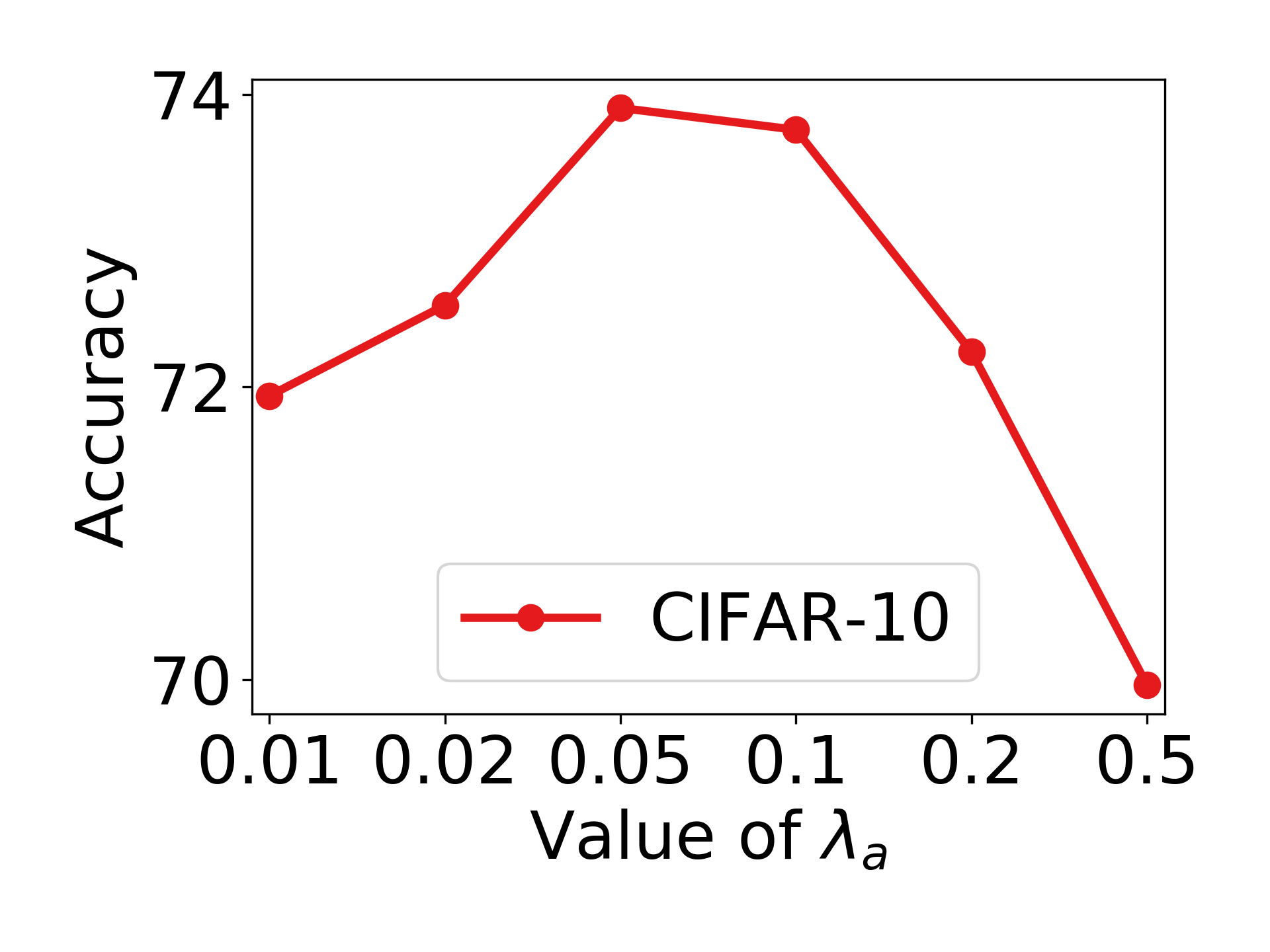}
\vskip -0.1in
\caption{Performance comparison of different choices of $\lambda_a$. Left: LeNet-5-HALF on MNIST. Right: AlexNet-HALF on CIFAR-10.}
\label{fig:lbda}
\vskip -0.1in
\end{figure}

We then evaluate the effect of different $\lambda_a$ values. We conduct experiments on LeNet-5-HALF and AlexNet-HALF with $\lambda_a=0.01,0.02,0.05,0.1,0.2,0.5$. The $\sigma$s are fixed to 1.5 and 2.0, respectively, as used in previous sections to achieve the best performance. Fig. \ref{fig:lbda} presents the performance of the student models with different values of $\lambda_a$. We see that for both configurations, when $\lambda_a=0.05$, the best test accuracies are achieved, i.e., 99.08\% and 73.91\% for LeNet-5-HALF and AlexNet-HALF, respectively. For $\lambda_a$ greater or smaller than 0.05, the test accuracies decrease. When a smaller $\lambda_a$ is introduced, neurons are usually not fully activated. On the other hand, for a larger $\lambda_a$, the activation values tend to be over-optimized, which in turn prevents the optimization of noise images' softmax outputs from getting close towards the soft targets.

\subsection{Effect of Each Component in the Loss Function}
In this section, we examine the effectiveness of the different components in the proposed approach to the performance of the student model. We evaluate whether modeling the shallower feature space helps improve the image synthesis process by comparing the performance of modeling the feature space of FC$_{-2}$ and the softmax space, respectively. Moreover, we also evaluate whether adding the extra activation loss helps improve the performance of the student model.

Tables \ref{tab:lenet_ablation} and \ref{tab:alex_ablation} report the performance with different components of the proposed approach on LeNet-5-HALF and AlexNet-HALF, respectively. It is observed that without modeling FC$_{-2}$ and the activation loss, i.e., modeling the softmax space with a multivariate normal distribution, accuracies of 98.92\% and 71.95\% are achieved with each configuration, respectively. When modeling the output space of the second last full connected layer (FC$_{-2}$), the performance are improved by 0.12\% and 1.59\% with LeNet-5-HALF and AlexNet-HALF, respectively. These improvements over their counterparts with ZSKD (98.77\% and 69.56\%), which models the softmax space with a Dirichlet distribution, validates the effectiveness of modeling the shallower feature space with a multivariate normal distribution. It can be observed that encouraging higher activation values also helps improve the performance, though the improvement is not as significant as that of modeling FC$_{-2}$ instead of the softmax space. With both FC$_{-2}$ and the activation loss implemented, the student models achieve the best performance.
\begin{table}[t]
\centering
\begin{tabular}{|C{1.8cm}|C{1.09cm}|C{1.09cm}|C{1.09cm}|C{1.09cm}|}
\hline
$\text{FC}_{-2}$ &  & \cmark &  & \cmark \\
\hline
Activation &  &  & \cmark & \cmark \\
\hline
Accuracy & 98.92\% & 99.04\% & 98.96\% & 99.08\% \\
\hline
\end{tabular}
\caption{Performance with different components of the proposed approach on LeNet-5 with the MNIST dataset.}
\label{tab:lenet_ablation}
\vskip -0.2in
\end{table}

\begin{table}[t]
\centering
\begin{tabular}{|C{1.8cm}|C{1.09cm}|C{1.09cm}|C{1.09cm}|C{1.09cm}|}
\hline
$\text{FC}_{-2}$ &  & \cmark &  & \cmark \\
\hline
Activation &  &  & \cmark & \cmark \\
\hline
Accuracy & 71.95\% & 73.54\% & 72.46\% & 73.91\% \\
\hline
\end{tabular}
\caption{Performance with different components of the proposed approach on AlexNet with the CIFAR-10 dataset.}
\label{tab:alex_ablation}
\end{table}

\begin{figure}[t]
\centering
\includegraphics[width=0.98\columnwidth]{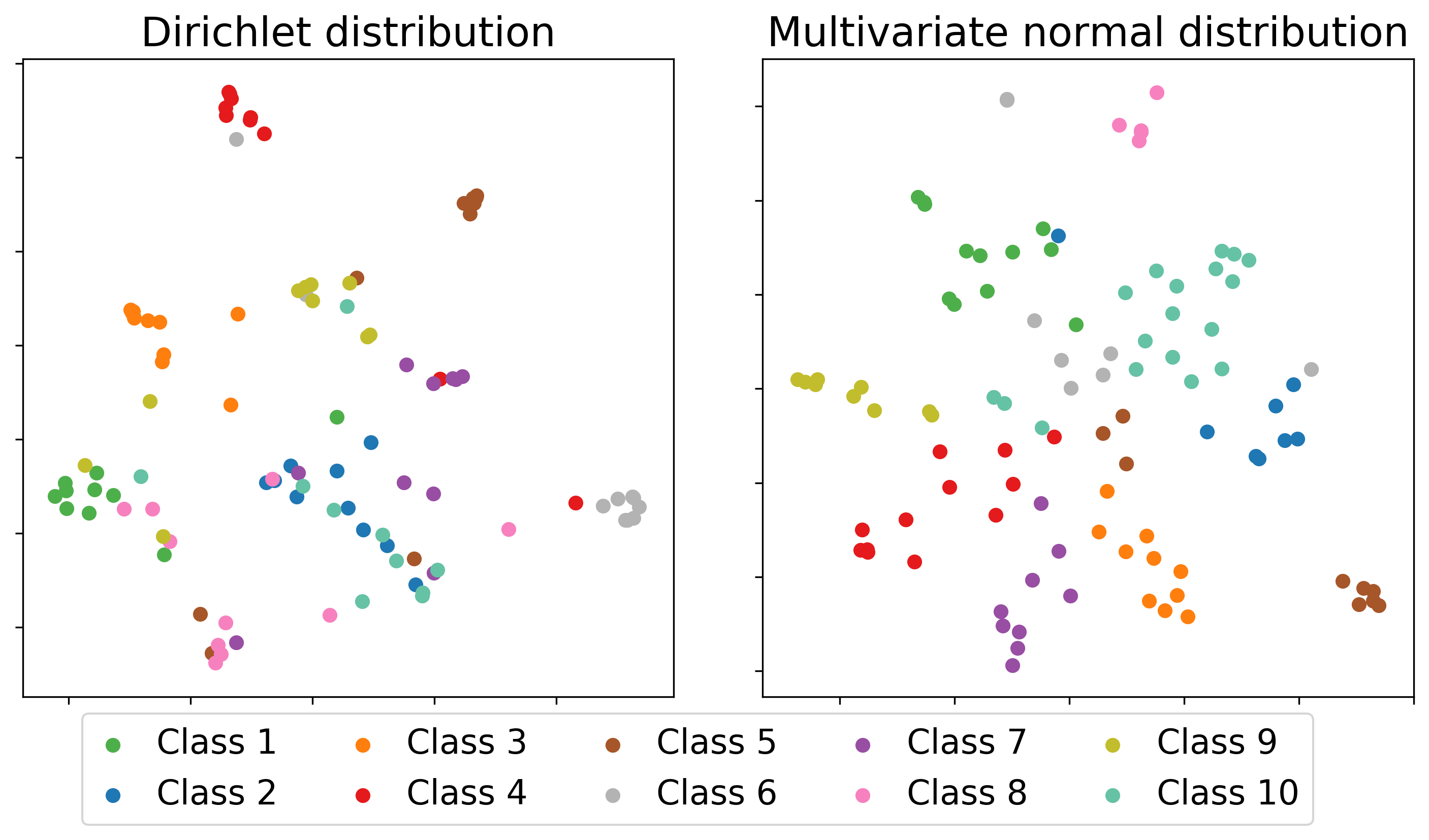}
\caption{Visualization of the 2-d embeddings of the generated soft targeted labels with ZSKD (Dirichlet distribution) and our proposed approach (multivariate normal distribution) via t-SNE. Figure best viewed in color.}
\label{fig:label_distribution}
\vskip -0.1in
\end{figure}

\subsection{Multivariate Normal vs. Dirichlet Distribution}
Since both ZSKD and our approach model the feature space with a prior probability distribution, we further investigate the differences between these two approaches in this sub-section. We generate 100 soft targeted labels using each approach with an AlexNet teacher model trained with CIFAR-10 for illustration. Fig. \ref{fig:label_distribution} gives the 2-d t-SNE \cite{maaten2008visualizing} embeddings of the soft targeted labels generated by ZSKD (with Dirichlet distribution) and our approach (with multivariate normal distribution). It is observed that samples generated from the multivariate normal distribution are clustered well, which are separable in the low-dimensional space. On the other hand, sampling from the Dirichlet distribution leads to a mixture of targets that belong to different classes in each cluster, which indicates that the generated labels are mismatched with its real category. This is because ZSKD generated labels for each category separately, and there is always a chance that the index corresponding to the maximal probability in the softmax output mismatches the real category. Actually, in our empirical study, Dirichlet distribution can produce around 20\% to 40\% labels that are mismatched, which substantially hurt the quality of the generated samples. On the other hand, modeling with a multivariate normal distribution considers the samples of all the classes as a whole, which can theoretically avoid the label mismatch problem.

\section{Conclusion}
\label{sec:con}
In this paper, we proposed a data-free knowledge distillation approach. We first modeled the intermediate feature space of the teacher model with a multivariate normal distribution and sampling from that distribution to generate soft targeted labels, which are then used to generate pseudo training samples as the transfer set. Finally, the student model is trained with the transfer set via a standard KD process. We evaluate the proposed approach with several benchmark architectures and datasets on the object classification task and the results demonstrate the effectiveness of our approach.

\bibliography{ref}
\end{document}